\documentclass[A4,12pt,conference]{smart_2025_Paper}

\usepackage{times}

\usepackage{graphicx}
\usepackage{amsmath}
\usepackage{amsfonts}
\usepackage{amssymb}
\usepackage[numbers]{natbib}
\usepackage{hyperref}
\usepackage{xcolor}
\usepackage[normalem]{ulem}         
\usepackage[detect-all]{siunitx} 
\usepackage{subcaption}
\usepackage{float}
\usepackage{booktabs} 

\definecolor{nblue}{rgb}{0.09, 0.768, 1}
\definecolor{caribbeangreen}{rgb}{0.0, 0.8, 0.6}

\newcommand\ayechsout{\bgroup\markoverwith{\textcolor{nblue}{\rule[0.5ex]{2pt}{0.8pt}}}\ULon} 
\def\sensor{\mathrm{S}}

\newcommand{\reals}{\mathbb R}
\newcommand{\xv}{\boldsymbol x}
\newcommand{\hv}{\boldsymbol h}
\newcommand{\Xv}{\mathbf X}
\newcommand{\yv}{\boldsymbol y}
\newcommand{\yvmean}{\left\langle \hat \yv \right\rangle}
\newcommand{\bv}{\boldsymbol b}
\newcommand{\zv}{\boldsymbol z}

\newcommand{\Wm}{\mathbf W}

\newcommand{\LE}{LE}
\newcommand{\rnn}{r}
\DeclareMathOperator{\mse}{MSE}

\newcommand{\tth}{t_{\rm th}}

\newcommand{\tshift}{t_{\rm shift}}

\newcommand{\Tsignal}{T}

\newcommand{\err}{e}

\title{Localization of Impacts on Thin-Walled Structures by Recurrent Neural Networks: End-to-end Learning from Real-World Data}

\author{Alexander Humer$^{\dag}$, Lukas Grasboeck$^{*,\dag}$, Ayech Benjeddou$^{\ddag}$}

\heading{Alexander Humer, Lukas Grasboeck and Ayech Benjeddou}

\address{
$^{\dag}$ Institute of Technical Mechanics, Johannes Kepler University Linz, Austria\\
e-mail: alexander.humer@jku.at
\and
$^{*}$Linz Center of Mechatronics GmbH, Linz, Austria\\
e-mail: lukas.grasboeck@lcm.at
\and
$^{\ddag}$
Institut Sup\'{e}rieur de M\'{e}canique de Paris, Saint Ouen, France \\
Universit\'{e} de Technologie de Compi\`{e}gne, Roberval, Compi\`{e}gne, France\\
e-mail: ayech.benjeddou@isae-supmeca.fr
}

\abstract{
    Today, machine learning is ubiquitous, and structural health monitoring (SHM) is no exception.
    Specifically, we address the problem of impact localization on shell-like structures, where knowledge of impact locations aids in assessing structural integrity.
    Impacts on thin-walled structures excite Lamb waves, which can be measured with piezoelectric sensors.
    Their dispersive characteristics make it difficult to detect and localize impacts by conventional methods.
    In the present contribution, we explore the localization of impacts using neural networks.
    In particular, we propose to use {recurrent neural networks} (RNNs) to estimate impact positions end-to-end, i.e., directly from {sequential sensor data}.
    We deal with comparatively long sequences of thousands of samples, since high sampling rate are needed to accurately capture elastic waves.
    For this reason, the proposed approach builds upon Gated Recurrent Units (GRUs), which are less prone to vanishing gradients as compared to conventional RNNs.
    Quality and quantity of data are crucial when training neural networks.
    Often, synthetic data is used, which inevitably introduces a reality gap.
    Here, by contrast, we train our networks using {physical data from experiments}, which requires automation to handle the large number of experiments needed.
    For this purpose, a {robot is used to drop steel balls} onto an {aluminum plate} equipped with {piezoceramic sensors}.
    Our results show remarkable accuracy in estimating impact positions, even with a comparatively small dataset.
}
\keywords{Structural Health Monitoring, Impact Localization, Guided Wave Propagation, Recurrent Neural Networks, Robot-Supported Experiments}

\begin{document}

\section{\MakeUppercase{Introduction}}
Impact localization on thin-walled structures plays a crucial role in structural health monitoring (SHM) and damage assessment, particularly in aerospace and automotive applications where lightweight and safety-critical components are common. 
Knowledge of impact locations is essential for assessing structural integrity and detecting potential damage in structures. 
While this problem has traditionally been approached using blends of physics and data-based methods, recent advances in machine learning have opened new possibilities for more effective solutions, see, e.g.,~\cite{vuquoc2023}.

Mechanical impacts on thin-walled structures generate guided elastic waves—commonly referred to as \emph{Lamb waves}~\cite{lamb1917}, which can be captured using a network of piezoelectric sensors.
Accurately determining the location of an impact in such structures, however, remains a challenging task due to the complex propagation characteristics of these waves. 
Lamb waves are highly dispersive and their behavior is further complicated in non-isotropic materials such as fiber-reinforced polymers, where anisotropy and inhomogeneity introduce additional complexity in wave propagation, see, e.g.,~\cite{ciampa2012,jang2015,xiao2024}.
Conventional approaches to impact localization predominantly rely on estimating the time-of-arrival (ToA) of the impact-induced wavefront at several distributed sensors.
Classical methods from the field of signal processing include, among several others, continuous wavelet transform, modified energy-ratio and the Akaike information criterion, see, e.g.,~\cite{grasboeck2023,grasboeck2024} for an overview.
These estimates are then used in triangulation-type algorithms to infer the impact location, see, e.g.,~\cite{Kundu_2014}.
While this methodology has been widely adopted, it typically utilizes only a small portion of the available signal information—namely, the wavefront arrival. 
The rich signal content that follows, including reflections at structural boundaries and mode conversions, is usually disregarded, despite its potential to enhance localization performance.

Approaches based on different kinds of neural networks can leverage a larger part of the information contained in sensor signals to accurately estimate impact positions. 
Recent contributions combine fully-connected networks for localization and recurrent structures to reconstruct force histories~\cite{huang2023}. 
Spatio-temporal graph convolutional networks for the localization of impacts on composite plates have been proposed in~\cite{zhao2023}. 
The framework of Gaussian processes is used in~\cite{xiao2025} to account for the inherently probabilistic nature of impact-induced waves in thin-walled structures. 

In this work, we propose an alternative approach that aims to exploit the full informational content of the sensor signals by learning to estimate the impact position directly from raw time-series data. 
In particular, we propose to use \emph{recurrent neural networks} (RNNs) to estimate impact positions \emph{end-to-end}, i.e., directly from \emph{sequential sensor data}. 
Our central hypothesis is that an end-to-end learning framework can yield more accurate and robust localization results, potentially with fewer sensors than those required in conventional methods.
Due to the high sampling rate needed to accurately capture Lamb waves, we deal with comparatively long sequences, which pose a significant challenge when training RNNs. 
Therefore, we employ more complex network architectures such as Gated Recurrent Units (GRUs)~\cite{cho2014}, which incorporate gating mechanisms to manage long-term dependencies more efficiently than standard RNNs.

Quality and quantity of data are crucial when training neural networks. 
Often, synthetic data is used, which inevitably introduces a reality gap. 
Here, by contrast, we train our networks using physical data from experiments, which requires automation to handle the large number of experiments needed. 
For this purpose, a collaborative robot is used to drop steel balls onto an aluminum plate equipped with piezoceramic sensors.

Our results show remarkable accuracy in estimating impact positions, even with a comparatively small dataset. 
This suggests that the proposed approach effectively leverages the rich information content in Lamb wave signals, potentially offering a more efficient and accurate alternative to conventional impact localization methods.

The present paper is organized as follows: Following this introduction, we briefly outline the network architecture of the proposed estimator. 
Section~\ref{sec:experiments} introduces the experimental setup for the fully automated generation of training data. 
The (minimal) data preperation and the training setup are described in Section~\ref{sec:training}.
Results are presented and discussed in Section~\ref{sec:results}, highlighting the model's performance across different sensor configurations and signal durations. 
Finally, Section~\ref{sec:conclusion} summarizes our findings and outlines directions for future work.

\section{\MakeUppercase{Methodology}}

\subsection{Impact location and sensor signals} 
The impacts on the plate occur in a target region $T \subset \mathbb R^2$.
The vector of true positions with respect to some fixed Cartesian frame is denoted by
    $\yv = \left( p_x , p_y \right)^T \in T$.
%
Let $S$ denote the number of sensors distributed across the structure to be monitored. 
We assume a fixed sampling rate, i.e., sensors take measurements at equidistant time points $t_i$, where $i=1, \ldots, L$. 
At each time $t_i$, sensors record signals $x_i^{(j)} \in D$, for $j=1, \ldots, S$, with $D$ being the range of sensor signals, which are then combined into a vector $x_i$:
\begin{equation}
    \xv_i = \xv(t_i) = \left( x^{(1)}(t_i) ,  \ldots , x^{(S)}(t_i) \right)^T \in D \times \ldots \times D \subset \mathbb R^S . 
\end{equation}
For notational convenience, sensor signals $\xv_i$ form the rows of a matrix $\Xv$,
\begin{equation}
    \Xv = \left( \xv_1 , \ldots , \xv_L \right)^T \in \reals^{L \times S} , 
\end{equation}
which represents the complete sensor data over time.

\subsection{Model architecture}
The proposed location estimator $\LE$ is supposed to predict the impact location $\yv$ from the (temporal) sequence of sensor signals $\Xv$.
In other words, $\LE$ maps the (spatio-)temporal patterns in the sensor data to a position in the target region $T$, which, from a mathematical point of view, translates into
\begin{equation}
    \LE : \reals^{L \times S} \to \reals^2 : \quad \Xv \mapsto \hat \yv = \LE (\Xv) ,
\end{equation}
where a hat is introduced for predicted positions.
Given the sequential nature of the sensor data $\Xv$, it is crucial to capture temporal dependencies  for accurate predictions. 
RNNs are a natural choice to represent such an estimator, since, owing to their sequential nature, they are capable of processing variable-length input sequences while maintaining a memory of prior time steps.
The memory of an RNN is represented by the hidden state $\hv_i$, which is computed from the previous hidden state $\hv_{i-1}$ and the current input $\xv_i$,
\begin{equation}
    \hv_i = \rnn (\hv_{i-1}, \zv_i) \in \reals^H , \qquad 
    \zv_i = \Wm^{\rm (in)} \xv_i + \bv^{\rm (in)} , \qquad
    \Wm^{\rm (in)} \in \reals^{H \times S} , \; \zv_i, \bv^{\rm (in)} \in \reals^H ,
\end{equation}
where $H$ denotes the dimensionality of the hidden state.
Note that we have introduced a linear layer to project the input vector $\xv_i$ onto a vector $\zv_i$, which has the same dimension as the hidden state.
Following machine learning conventions, $\Wm^{\rm (in)}$ is referred to as weight-matrix and $\bv^{\rm (in)}$ denotes the bias vector.

Intuitively, the prediction for the impact location is made once the whole sequence of sensor data has been processed, in our case, by a linear map applied to the final hidden state $\hv_L$:
\begin{equation}
    \hat \yv = \Wm^{\rm (out)} \hv_L + \bv^{\rm (out)}, \qquad \Wm^{\rm (out)} \in \reals^{2 \times H} , \bv^{\rm (out)} \in \reals^2 .
\end{equation} 

The specific RNN architecture is subject to practical constraints that arise from the particular SHM application of impact monitoring.
Lamb waves propagate at comparatively high velocities.
For the typical frequencies excited by impacts, the phase velocities of the fundamental anti-symmetric ($A_0$) mode are \qty{1000}{\meter \per \second} or higher for the aluminum panel used in our experimental study; see, e.g., \cite{grasboeck2024}.
For this reason, high sampling rates are required to capture the propagation of Lamb waves, which implies long sequences of sensor data.

The long input sequences resulting from high sampling rates challenge standard RNNs, which are known to suffer from vanishing gradients when modeling long-term dependencies.
GRUs, in contrast, introduce gating mechanisms that control the flow of information to mitigate the problem of vanishing gradients, while maintaining a lower complexity than Long Short-Term Memory (LSTM)~\cite{hochreiter1997}.


\subsection{Loss function and training}
The estimator is trained in a supervised manner by minimizing the mean squared error (MSE) between predicted positions $\hat \yv$ and true positions $\yv$, which serves as natural choice for the loss function.
Our initial experiments, however, have shown that the estimator is strongly conditioned to the length of the input sequence if the prediction is made only after the entire sequence has been processed.
When training is performed with a fixed sequence length $L$, the model learns to make accurate predictions only after exactly $L$ time steps, but neither for longer nor for shorter sequences.
To address this issue, we modify the training procedure such that the estimator makes intermediate predictions at each time step $t_i$.
By minimizing the loss not only at the end of the sequence, but for every time step, i.e.,
\begin{equation}
    \mse = \frac 1 L \sum_{i=1}^L \left\Vert \hat \yv_i - \yv \right\Vert^2 , 
\end{equation}
the model learns to generalize across variable-length inputs.
Using the above loss, we force the estimator to make predictions for every sensor data $\xv_i$ being processed, so that the length of the training sequence is no longer ``learned''.
This strategy is particularly natural in online monitoring systems, which are meant to provide location estimates continuously, i.e., sample by sample, as new data arrives.
That raises the question, of course, what the model predicts prior to the arrival of impact-induced waves at the sensors, i.e., before information on the location of their source becomes is contained in the sensor signals.
Initially, the prediction defaults to a mean estimation which reflects the distribution of impact locations in the training data, e.g., the center of the plate if impacts are uniformly distributed.

\section{\MakeUppercase{Experimental setup}}
\label{sec:experiments}
As with any data-driven approach, both the quality and quantity of data are crucial for creating accurate and reliable estimators.
Due to the limited availability of real-world data, it has become common practice—particularly in the AI domain—to use synthetic data, especially when training neural networks, which are notorious for their high requirements in terms of data.
In physics-related problems, numerical simulation provides effective means to generate synthetic training data, with real-world experiments typically requiring significantly greater expenses.
In the present application, however, the situation is different.
Owing to the fine temporal and spatial resolution required, simulations of transient Lamb waves that propagate in thin-walled structures are prohibitively demanding in terms of computational efforts to create a reasonable amount of training data.
For this reason, physical experiments are key to the realization of the proposed impact localization estimator. 
As a side effect, we avoid issues related to what is referred to as \emph{reality gap}, i.e., an inevitable modeling error in simulation models.

\begin{figure}[ht!]
    \centering
    \includegraphics[width=0.9\textwidth]{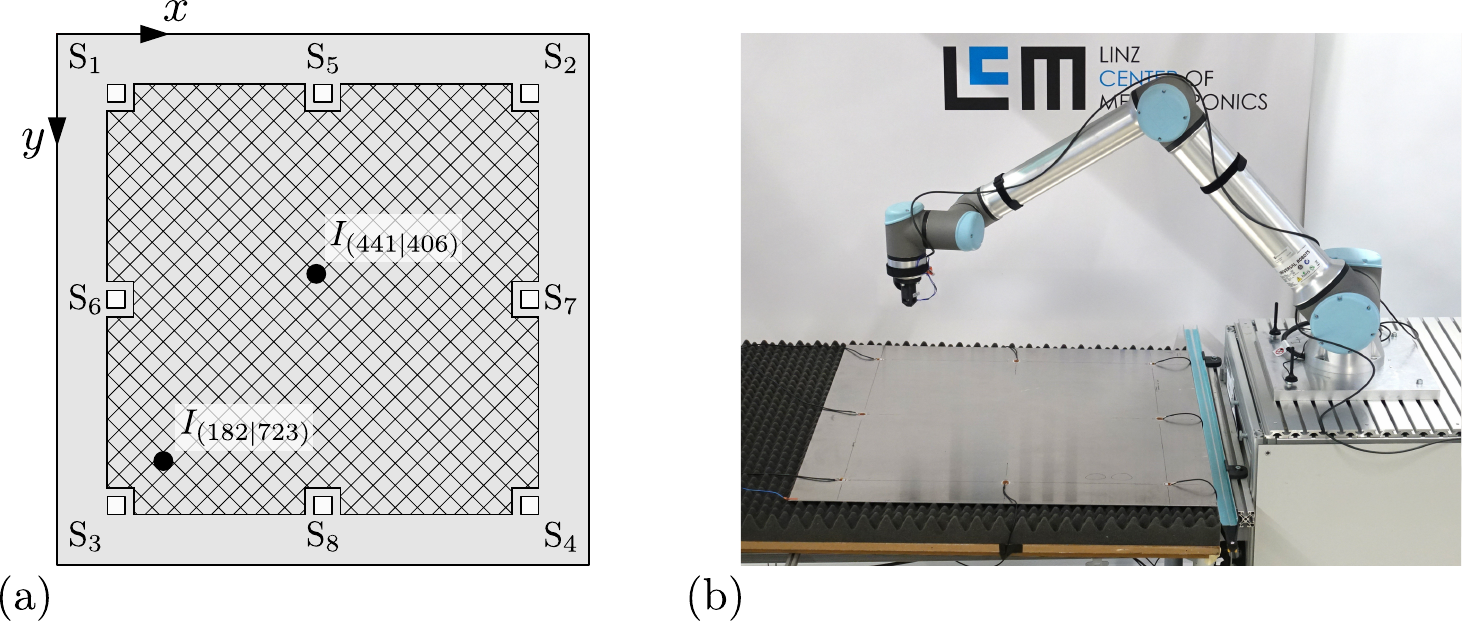}
    \caption{Experimental setup for impact localization on an aluminum plate:
    Schematic representation of the plate with the position of the sensors $\sensor_j$, $j=1,\ldots,8$ (a).
    The crosshatched area indicates the region where impacts are applied.
    Two example impact positions from the dataset of \num{5000} impacts are marked as black dots, with their $(x,y)$-coordinates given in millimeters in parentheses.
    Picture of the robotic impact test setup used to generate the dataset for training neural networks (b).}
    \label{fig:exp_setup}
\end{figure}

In the present work, a nearly square aluminum plate measuring $x \times y \times z = \qtyproduct{904 x 902 x 2}{\milli\meter}$ serves as an example of a thin-walled structure to be impacted.
To capture the propagation of impact-induced waves, eight piezoelectric patch transducers (PI Ceramic P-876.SP1) denoted by $\sensor_i$, with $i=1,\ldots,8$, are bonded to the plate's top surface in a distance of \qty{100}{\milli\meter} from the edges, see Fig.~\ref{fig:exp_setup} (a).
Sensors $\sensor_1$ -- $\sensor_4$ are placed at the corners of the plate; $\sensor_5$ -- $\sensor_8$ are located at the centers of the plate's edges.
The center positions of the sensors with respect to the global Cartesian frame placed in the upper left corner are listed in Tab.~\ref{tab:sensor_positions}.
The active area of the patches is \qtyproduct{10 x 10}{\milli\meter} in size.
\begin{table}[ht!]
    \small
    \centering
    \caption{Positions of the sensors on the aluminum plate. The coordinates are referring to the centers of the sensor patches.}
    \label{tab:sensor_positions}
    \begin{tabular}{*{9}{c}}
    \toprule
    Coordinates & $\sensor_1$ & $\sensor_2$ & $\sensor_3$ & $\sensor_4$ & $\sensor_5$ & $\sensor_6$ & $\sensor_7$ & $\sensor_8$ \\
    \midrule
    $x$ / \si{\milli\metre} & \num{100} & \num{803} & \num{100} & \num{803} & \num{452} & \num{100} & \num{803} & \num{452} \\
    $y$ / \si{\milli\metre} & \num{100} & \num{100} & \num{802} & \num{802} & \num{100} & \num{451} & \num{451} & \num{802} \\    
    \bottomrule
    \end{tabular}
\end{table}
We want to avoid impacts on or directly next to the sensors, which is why we leave a margin of \qty{5}{\milli\meter}.
The crosshatched region marks the area of potential impacts on the plate.
The sensor signals are sampled at \qty{1}{\mega\hertz} using a National Instruments NI USB-6366 measurement system, which digitizes voltages in the range of \qty{+-10}{\volt} through a \num{16}-bit analog-to-digital converter, corresponding to an amplitude resolution of \qty{0.31}{\milli\volt}.
To protect the data acquisition system from high voltages, a 1:10 probe is used.

To generate a sufficiently large and diverse dataset for training the estimator, a large number of impact events must be recorded. 
Manual generation of impacts is impractical and prone to inaccuracies in positioning. 
Therefore, a robotic system is employed to automate the experiments, ensuring precise and repeatable impact locations and enabling efficient collection of high-quality data as needed for training the estimator.
For this purpose, a collaborative robot (Universal Robots UR10) is employed, see Fig.~\ref{fig:exp_setup} (b). 
An end effector equipped with an electromagnet picks up a steel ball from a designated position and releases it at the precise impact location on the plate. 
The plate is supported from the bottom by a slightly inclined foam mat (less than \SI{1}{\degree}), ensuring that, after impact, the ball rolls into a guide rail that returns it to the pickup position, where a magnetic proximity sensor verifies the ball's return before initiating the next drop. 
The robot operates through a real-time data exchange (RTDE) interface \cite{UR_RTDE_Guide}. 

\section{\MakeUppercase{Data preparation and training}}
\label{sec:training}
\begin{figure}[ht!]
    \centering
    \includegraphics[width=\textwidth]{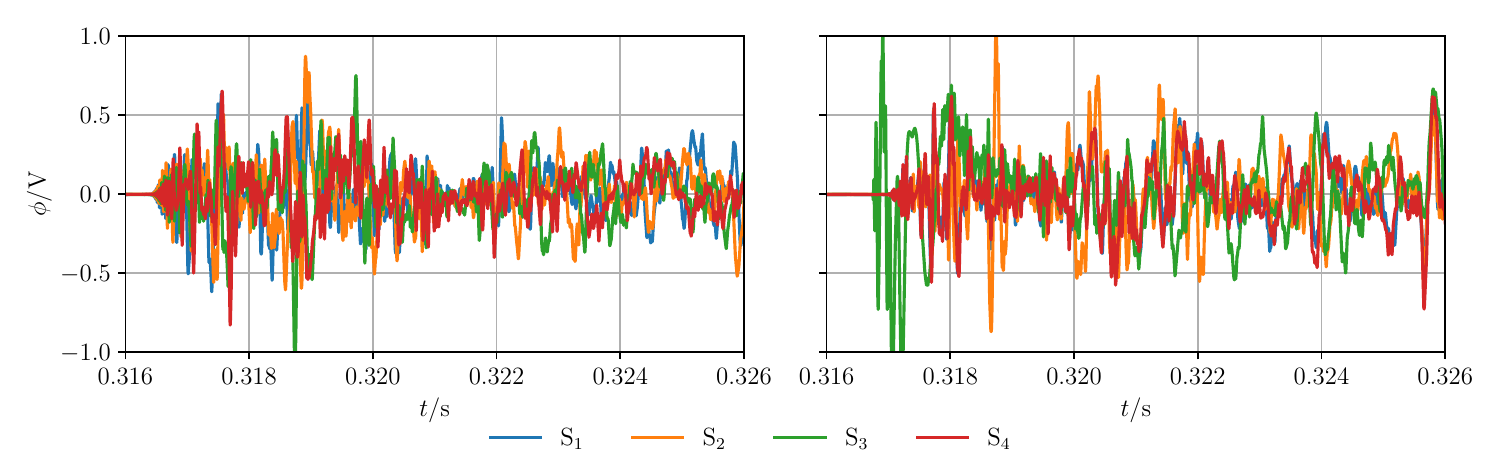}
    \caption{Representative sensor signals:
    Impacts at a centered position (\qty{441}{\milli\metre}, \SI{406}{\milli\metre}) (left) and
    close to a corner (\qty{182}{\milli\metre}, \SI{723}{\milli\metre}) (right).}
    \label{fig:selected_sensor_signals}
\end{figure}

Within the present work, balls are dropped from a fixed height of \qty{200}{\milli\meter} onto \num{5000} random positions sampled from a uniform distribution.
Figure~\ref{fig:selected_sensor_signals} shows two representative impacts and the corresponding signals of the corner sensors $\sensor_1$ -- $\sensor_4$.
The locations of the impacts are shown in Fig.~\ref{fig:exp_setup}. 
For a rather centered impact (\qty{441}{\milli\meter}, \qty{406}{\milli\meter}), sensor signals are relatively similar from a qualitative point of view. 
The waves excited by the impact arrive at almost the same time at the four sensors.
For an impact close to a corner (\qty{182}{\milli\meter}, \qty{723}{\milli\meter}), the individual sensor signals differ significantly from one another. 
Unsurprisingly, the sensor closest to the impact ($\sensor_3$) records waves earlier than the other three sensors. 
We also note a higher amplitude, since, due to the shorter distance, the waves attenuate less. 
The peak voltages $\phi$ are mostly in the range of \qty{+-1}{\volt}; the maximum voltage occurring in the dataset is \qty{2.21}{\volt}. 
The noise level is three orders of magnitude smaller; we observe constant offsets in the sensor signals of up to \qty{3}{\milli\volt}.
Except for a downsampling to \qty{250}{\kilo\hertz}, no further processing is applied to the sensor signals.

Measurements are triggered by the robot, which turns off the magnet to drop the steel ball as soon as it reaches the desired position.
Between the trigger and the impact on the plate, we see a consistent delay of approximately \qty{316}{\milli\second}.
For efficient training, we do not want to include the entire non-informative pre-impact data. 
To avoid conditioning the network to a fixed arrival time, we apply a random cropping strategy around the detected signal onset.
For each impact, we determine the time $\tth$ at which the signal level rises beyond a threshold of \qty{50}{\milli\volt}.
We then crop a window of length $\Tsignal$ from the sensor signals, i.e.,  $t \in \left[\tth - \tshift , \tth - \tshift + T \right]$, where $\tshift$ is randomly chosen from the interval $\tshift \in \left[ \qty{0.5}{\milli \second} , \qty{2}{\milli \second} \right]$.

The target positions to be predicted by the estimator are provided in \unit{\meter}, which implies that numerical values range from \numrange{0.1}{0.8}.
No further normalization is applied in what follows.
The dataset comprises \num{4000} impacts for training, \num{500} for validation (hyperparameter tuning, early stopping) and \num{500} for testing purposes.
The proposed model is implemented in PyTorch~\cite{pytorch} and trained on an NVIDIA A30 GPU.


\section{\MakeUppercase{Results}}
\label{sec:results}

The following results aim to demonstrate the feasibility of the proposed end-to-end learning approach for impact localization, rather than to achieve optimal performance through extensive hyperparameter tuning.
To this end, we use a baseline model configuration based on empirical estimates that have proven effective in our training experiments.
Specifically, the estimator consists of a single GRU layer with an input and hidden size of $H = 32$ ($H = 64$ when using eight sensors).
Training is conducted over 5000 episodes using the AdamW optimizer with an initial learning rate of \num{1e-2} and a step-wise learning rate decay by a factor of \num{0.9} every \num{100} episodes.
A fixed batch size of \num{500} is (mostly) used throughout training; batches are sampled randomly from the training set.
Depending on the length $T$ of the crop window, which ranges from \qtyrange{5}{20}{\milli\second}, the length of the input sequence varies from \numrange{1250}{5000} samples, which is a considerable length in the context of RNNs. 

\begin{figure}[ht!]
    \centering
    \includegraphics[width=\textwidth]{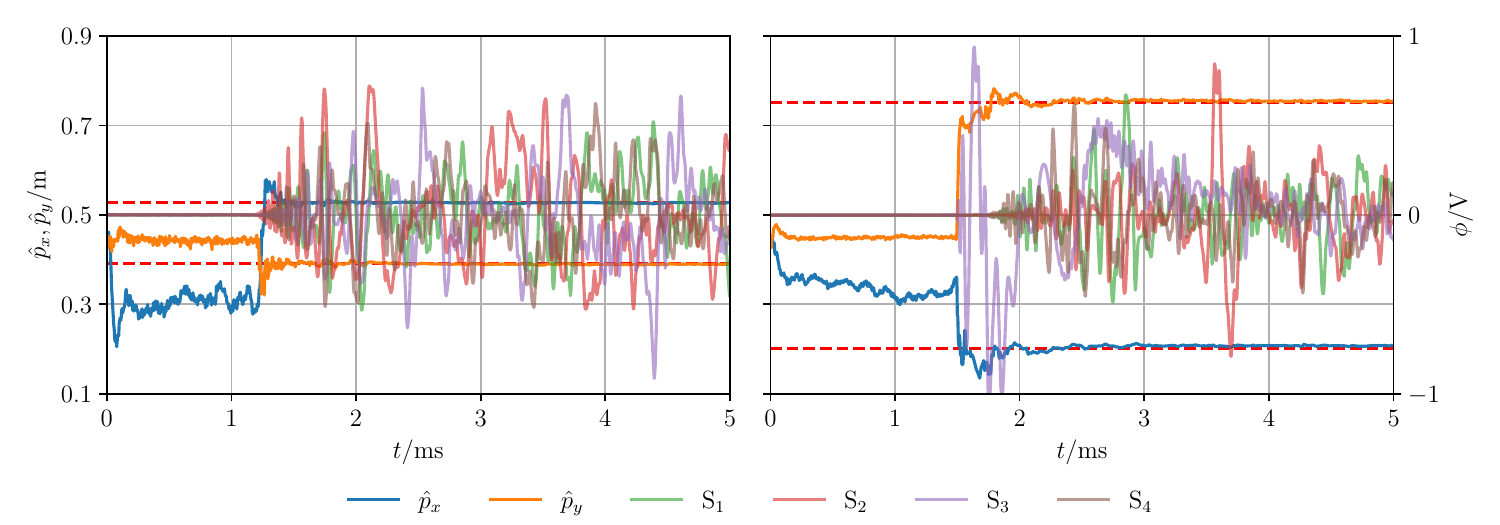}
    \caption{
        Sequential predictions of impact locations ($S=4$, $T = \qty{20}{\milli\second}$) for impacts at (\qty{527}{\milli\meter}, \qty{391}{\milli\meter}) (left) and (\qty{200}{\milli\meter}, \qty{750}{\milli\meter}) (right): Predicted positions $\hat p_x$, $\hat p_y$ are illustrated along with corresponding input sensor signals $\sensor_1, \ldots, \sensor_4$. As long as sensor signals do not contain information on impacts, no meaningful estimates can be made. As impact-induced waves arrive at the sensors, estimated impact locations converge to (and remain at) true positions (red dashed lines).
    }
    \label{fig:prediction}
\end{figure}

Figure~\ref{fig:prediction} illustrates the evolution of predicted impact locations $\hat \yv$ over \qty{5}{\milli\second} as the sensor signals are sequentially processed for two positions of the test set, i.e., (\qty{527}{\milli\meter}, \qty{391}{\milli\meter}) (left) and (\qty{200}{\milli\meter}, \qty{750}{\milli\meter}) (right), using an estimator trained on $T = \qty{20}{\milli\second}$ sequences recorded by four sensors $\sensor_1, \ldots, \sensor_4$.
Initially, the sensor signals do not contain any information related to an impact, which is why the predictions fluctuate around seemingly random values. 
As soon as impact-induced waves arrive at the sensor, the predictions quickly converge to the true positions, which are indicated by red dashed lines, and remain almost constant subsequently.
For the second, off-center impact (right), we can observe that the estimator learns to predict impact locations using a single sensor only. 
Although a sensor signal is only present in $\sensor_3$ at first, the estimated impact position is already close to the true position before the remaining sensors begin to measure.

Being computed over the entire input sequence, the loss is not an appropriate measure of the estimator's prediction quality. 
As an intuitive performance metric, we therefore compute the Euclidean distance between the true impact location $\yv$ and the predicted location $\hat \yv$, which is averaged over the last $N = \num{10}$ time steps:
\begin{equation}
    \err = \left\Vert \yvmean - \yv \right\Vert , \qquad
    \yvmean = \frac 1 N \sum_{i=L-N}^L \hat \yv_i .
    \label{eq:distance}
\end{equation}

\begin{figure}[ht!]
    \centering
    \includegraphics[width=\linewidth]{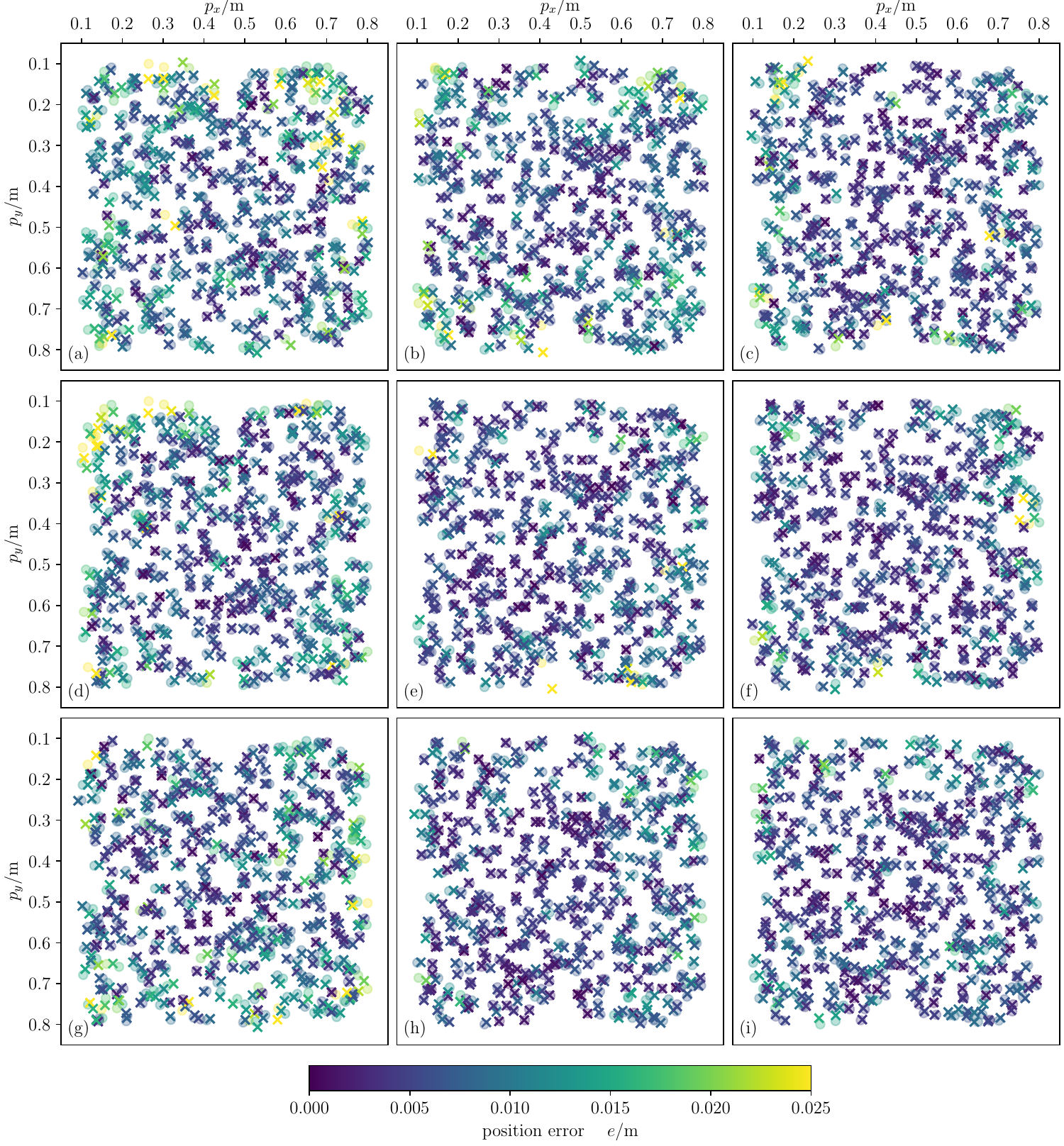}
    \caption{
        Impact localization accuracy (test set): True positions are indicated by semi-transparent round markers, estimated positions by crosses. Colors respresent the Euclidean distance between true and estimated positions. Rows correspond to the number of sensors used for the prediction ($S=2,4,8$); columns represent different lengths of time windows ($T=\qtylist[list-separator={,},list-final-separator={,}, list-units=single]{5;10;20}{\milli\second}$).
    }
    \label{fig:results}
\end{figure}

%
%
Figure~\ref{fig:results} illustrates the accuracy of our proposed estimator across all impacts in the test set. 
The true positions of impacts are represented by semi-transparent round markers, while the estimated positions are denoted by crosses. 
The color coding indicates the Euclidean distance between true and estimated positions, cf.~Eq.~~\eqref{eq:distance}, providing a visual representation of the distribution of the estimation error over the plate structure.
For the sake of clarity, the upper limit of the colorbar is set to a value of \qty{25}{\milli\meter} rather than the maximum error in order to emphasize nuances in the variation of the prediction quality.
The results are organized in a table where rows correspond to different numbers and configurations of sensors used for prediction, while columns represent different time window lengths, i.e., $T=\qtylist[list-separator={,},list-final-separator={,}, list-units=single]{5;10;20}{\milli\second}$. 
Specifically, rows in Fig.~\ref{fig:results} correspond to: two adjacent sensors $\sensor_1, \sensor_2$ in the top row (a)--(c), the four corner sensors $\sensor_1$ -- $\sensor_4$ in the middle row (d)--(f), and all eight sensors $\sensor_1$ -- $\sensor_8$ in the bottom row (g)--(i). 

When using all eight sensors, the dimensionality of the input per time step increases accordingly.
Therefore, the input and hidden size of the GRU cell are both increased to $H = \num{64}$ to account for the larger amount of data that is being processed. 
For the longest time window ($T = \qty{20}{\milli\second}$), the batch size then has to be reduced for GPU memory constraints; for short windows ($T = \qty{5}{\milli\second}$), we increase the batch size to improve convergence and mitigate overfitting.

\begin{table}[ht!]
    \sisetup{range-phrase=\,--\,} 
    \small
    \centering
    \caption{
        Impact localization accuracy: Mean error and standard deviation of the position error $e$ for different lengths of the input signals and different numbers of sensors (Note: \textbf{bold} values indicate variations from the base model architecture and training setup).
    }
    \label{tab:results}
    \begin{tabular}{c|ccccc|cc}
    \toprule
        & Sensors  & $T / \si{\milli\second}$ & $L$ & $H$ & $B$ & $\mathrm{mean} (e) / \si{\milli\meter}$ & $\mathrm{std} (e) / \si{\milli\meter}$\\
    \midrule 
    (a) & \numrange{1}{2} & \num{ 5} & \num{1250} & \num{32} & \num{500} & \num{9.146} & \num{5.845} \\
	(b) & \numrange{1}{2} & \num{10} & \num{2500} & \num{32} & \num{500} & \num{7.808} & \num{5.445} \\
	(c) & \numrange{1}{2} & \num{20} & \num{5000} & \num{32} & \num{500} & \num{6.417} & \num{4.945} \\
    \midrule
	(d) & \numrange{1}{4} & \num{ 5} & \num{1250} & \num{32} & \num{500} & \num{8.029} & \num{5.263} \\
	(e) & \numrange{1}{4} & \num{10} & \num{2500} & \num{32} & \num{500} & \num{6.110} & \num{4.941} \\
	(f) & \numrange{1}{4} & \num{20} & \num{5000} & \num{32} & \num{500} & \num{5.842} & \num{3.941} \\
    \midrule
	(g) & \numrange{1}{8} & \num{ 5} & \num{1250} & \bf \num{64} & \bf \num{1000} & \num{8.110} & \num{5.385} \\
	(h) & \numrange{1}{8} & \num{10} & \num{2500} & \bf \num{64} & \num{500} & \num{5.902} & \num{3.633} \\
	(i) & \numrange{1}{8} & \num{20} & \num{5000} & \bf \num{64} & \bf \num{250} & \num{5.808} & \num{3.497} \\
    \bottomrule
    \end{tabular}
\end{table}
The mean prediction error and its standard deviation over the test set are listed in Tab.~\ref{tab:results}.
Our results reveal a clear and consistent trend: longer input windows generally lead to better localization accuracy. 
This suggests that the model benefits from later parts of the sensor signals, such as reflections and features related to the dispersive nature of Lamb waves, rather than relying solely on the initial wave front. 
Additionally, using more sensors improves performance, although the gains from increasing the number of sensors from four to eight are relatively modest. 
Interestingly, even with just two sensors, the estimator is capable of delivering accurate predictions, which is something traditional triangulation approaches typically cannot achieve, since they usually require three sensors at least.
We emphasize that the mean position errors are smaller than the dimensions of the piezoelectric patches. 

\section{\MakeUppercase{Conclusion and Outlook}}
\label{sec:conclusion}
We presented an end-to-end approach for localizing impacts on thin-walled structures using GRU-based recurrent neural networks. 
Unlike conventional methods that rely on time-of-arrival estimation, our model learns directly from raw sensor signals and captures both early and late-arriving wave components. 
Trained entirely on experimental data collected through a robot-assisted setup, the method achieves high localization accuracy—even with a limited dataset and as few as two sensors.

Future work will focus on extending the method to more complex structural scenarios, including composite shells and reinforced structures. 
We also plan to investigate optimal sensor placement strategies and sensor shapes. 
Further, we want to explore more advanced network architectures, such as transformers, to further enhance performance of end-to-end impact localization.

\bibliographystyle{IEEEtranN}
\bibliography{References}
\end{document}